# An Ontology for Satellite Databases

Robert J. ROVETTO[a, b, 1,2]

[a,] Research Affiliate, Center for Orbital Debris Education and Research (CODER), University of Maryland
[b] National Aeronautics and Space Administration (NASA) Datanauts, 2017

**Abstract.** This paper demonstrates the development of ontology for satellite databases. First, I create a computational ontology for the Union of Concerned Scientists (UCS) Satellite Database (UCSSD for short), called the UCS Satellite Ontology (or UCSSO). Second, in developing UCSSO I show that The Space Situational Awareness Ontology (SSAO)-—an existing space domain reference ontology—-and related ontology work by the author (Rovetto 2015, 2016) can be used either (i) with a database-specific local ontology such as UCSSO, or (ii) in its stead. In case (i), local ontologies such as UCSSO can reuse SSAO terms, perform term mappings, or extend it. In case (ii), the author_s orbital space ontology work, such as the SSAO, is usable by the UCSSD and organizations with other space object catalogs, as a reference ontology suite providing a common semantically-rich domain model. The SSAO, UCSSO, and the broader Orbital Space Environment Domain Ontology project is online at https://purl.org/space-ontology and GitHub. This ontology effort aims, in part, to provide accurate formal representations of the domain for various applications. Ontology engineering has the potential to facilitate the sharing and integration of satellite data from federated databases and sensors for safer spaceflight.

## 1. Introduction

This paper demonstrates the development of ontology for satellite databases. First, I develop a computational ontology for the Union of Concerned Scientists (UCS) Satellite Database (Union of Concerned Scientists (UCS) [1]) (UCSSD for short), called the UCS Satellite Ontology (or UCSSO). Second, in developing the UCSSO I show that the Space Situational Awareness Ontology (SSAO) [5](Rovetto and Kelso 2016)—an existing domain reference ontology—and related ontology work by the author [9,8,6,5,4] can be used either (i) with a database-specific local ontology such as UCSSO, or (ii) in its stead. In case (i), ontologies such as UCSSO can reuse SSAO terms, or map their own terms to it. In case (ii), the author_s space domain ontology work is usable by the UCSSD and organizations with other space object catalogs as a general reference ontology providing a common semantically-rich domain model. The SSAO, UCSSO, and the broader Orbital Space Environment Domain Ontology project is online via http://purl.org/space-ontology./[3].

With further development, this ongoing work serves as a case study and proof-of-concept for ontology engineering for space situational awareness (SSA) data in general, and for the orbital space ontology project (Rovetto 2016a). The aim is not simply to represent and conceptually analyze astrodynamic, astronautical and SSA entities in a formal and high-level manner. The project aims also to better manage, represent and reason over space data; facilitate knowledge sharing; improve SSA for safer spaceflight, and determine if ontology engineering can in fact do so. It should therefore be of interest to astroinformaticists [17](Borne 2010), SSA professionals, satellite operators, database administrators, philosophers, ontology engineers and computer scientists.

Specific goals for these ontologies are at least two-fold. One is to represent the realities of the domain: satellites, other space objects, their interactions and environments, orbits, etc. Two, to facilitate data exchange, integration, search; knowledge modeling, and semantic interoperability among federated space object databases and sensors. Ontology-based data queries, for instance, involve searches for satellites satisfying certain criteria. Potential answers may yield useful or novel information about a particular satellite, or satellite populations, their behavior, and orbital characteristics.

A *computational ontology* [13][15] consists of a set of defined class and relationship terms that are given a formally specified semantics. Ontologies represent the content and structure of a subject matter (domain)[14], focus on meaning of that content (and data), and are intended to communicate some understanding, common knowledge or conceptualization of that domain. Consider general knowledge of astrodynamics shared among satellite operators, as well as generic categories/concepts such as Satellite, Orbit, Inclination, etc. A *taxonomy,* by contrast, consists of a set of undefined terms typically structured according to a child-parent or class-subclass relation. Ontologies have taxonomies as proper parts, but take the extra step to make implicit meaning (such as definitions and assumptions) formally explicit. This affords the specification of various relations between classes to more accurately represent the actual relationships between domain objects. In short, "[o]ntologies provide […] terms, their meanings, their relations and constraints, etc." to model a domain [16]. Another goal is therefore the development of coherent and accurate space vocabularies.

---

[1] Emails: rrovetto@terpalum.umd.edu (ontologos@yahoo.com)
[2] This paper and the author's referenced papers are independent work, not undertaken with or at the author's current or past affiliations.
[3] Orbital Space Domain Ontology project landing page: http://purl.org/space-ontology.
UCSSO is at http://purl.org/space-ontology/ucsso/. The SSAO is at http://purl.org/space-ontology/ssao.

*Database schemas*, by contrast, define the structure and constraints of data/databases [14], focusing on data rather than what the data is about. Meaning for schemas is captured in data-dictionaries, normally separate from the schema, but which rarely change as the database changes. Ontologies, by contrast, change as databases do and as new knowledge is discovered. In other words, the information in a data-dictionary is effectively integrated into the ontology resulting in both a human and machine readable artifact. Ontologies and databases focus on general categories/classes and instances (of classes), respectively. The UCSSD, for example, stores instance data on over 1000 operational satellites, and "[t]he data in the UCS Satellite Database come from public sources, with much of the information provided by the satellite owners themselves"[2]. Ontologies adopt the open world assumption, whereas schemas adopt the closed world assumption—this is important because queries may yield different answers. Both employ knowledge representation languages, such as Common Logic Interchange Format (CLIF)[19], and the Web Ontology Language (OWL)[20]. Ontologies can require more computing power depending on their complexity and can present challenges with scalability [21], but they provide a richer representation of the domain and the data. Additionally, ontologies effectively reduce the programming complexity by effectively taking the domain model out of code. With an ontology-based information system, users are afforded a means for knowledge discovery and decision-support. Table 1 summarizes key features of ontologies.

| **Common Domain or Application Vocabulary:** | A system of interrelated categories: general but domain-specific terms | | |
|---|---|---|---|
| **Structured Terminology:** | Class-subsumption & other relations between category terms | | |
| **Formally defined terms:** | Natural language definitions | Artificial language definitions (ontology language, e.g., CLIF, OWL) | Rules, Logical Axioms |
| **Presents a domain or application model:** | Common knowledge model of the domain | Expresses meaning. Represents real-world domain phenomena. | |
| **Scalability & Reusability:** | Other ontologies can import or extend | Federated databases can use with their own ontologies, or as theirs (i & ii, above) | |
| **Analysis** | Reason over, and query the information (e.g., SPARQL, SWRL, DL Query, etc.) | | |

**Table 1.** Some Central features of an ontology

The next section describes the UCSSO taxonomy, terms of which correspond to both UCSSD terms [3] and domain-specific categories. UCSSD terms along with their matching ontology classes are discussed. Section 3 introduces the SSA Ontology, the Orbital Debris Ontology (ODO), the space ontology project of which they are a part; demonstrates the overlap with UCSSO. UCSSO classes are either found within the SSAO (and related ontologies by the author), or can be mapped to them. The SSAO is therefore offered as a domain ontology for the UCSSD and other satellite databases in the SSA and astronomy communities. Section 4 points to content in the domain that is in need of formalization, and 5 mentions future work.

## 2. Translating Database Terms to Ontology Terms

The UCSSD houses information about actual satellites: their names, national origins, orbital data, sources of data, etc. One option for ontological categories and characterizations of this information are as follows: satellite names are types of identifiers, national origins are social or political features, and orbital characteristics are physical and/or geometric properties (relational or otherwise).

Table 2 lists the field terms from the UCSSD file [3], their corresponding ontology classes, and a description/comment. Classes are implicitly structured with the *is a* class-subclass relation, are camel-cased with underscores separating words and occasionally in bold in the main text of the paper. Indentation signifies class-*subsumption*. Relational terms are italicized in text with underscores as well.[4] In conjunction with a formalized semantics (definitions, constraints, logical axioms, rules, etc.), they are intended to represent the kinds and relationships found in the real-world domain of artificial satellites. Definitions are drawn from the UCSSD Manual [22] where possible, and satellite and astrodynamics literature, space thesauri, or other references as needed.

---
[4] Relational terms are called Object Properties or Data Properties in the Protégé ontology editor, the latter of which takes some alpha-numerical value as one relatum or argument. Non-relational categories are called classes.

Some database terms are resolved into one or more classes and related with a relational term. This more precisely captures the domain reality by making ontological and conceptual distinctions explicit. Note that different taxonomies and classifications are possible. I provide one option. No claims to completeness are made, and the ontologies described or mentioned in this communication are works in progress. Figure 1 displays all classes using the OntoGraf plugin in the Protégé ontology editor [23]. Finally, although I do not do so here, formalizing supplemental information from the UCSSD comment field should make the ontology richer because it contains partonomic and other details of the individual satellites.

| DATABASE TERM/FIELD | ONTOLOGY TERM (CLASS) | COMMENT / DESCRIPTION |
|---|---|---|
| Name of Satellite, Alternate Names | | Resolved into two classes |
| | Satellite_Name | The current, primary, name of the satellite. |
| | Alternate_Satellite_Name | Past names, or synonyms. |
| Country/Org of UN Registry | Resolved into Country and Organization classes, and the following relations: *is_registered_country_in_UN_Register_of_Space_Objects* and *is_registered_organization_in_UN_Register_of_Space_Objects* | |
| Country of Operator/Owner | Resolved into classes: Country, Operator, Owner; and relations: *has_Operator*, *has_Owner*, and *has_Country_of_Origin* | |
| Operator/Owner | | Resolved into two ontology classes |
| | Operator | The operator of the satellite |
| | Owner | The owner of the satellite |
| Users | User<br>   Civil_User<br>      Academic_User<br>      Amateur_User<br>   Commercial_User<br>   Government_User<br>   Military_User | Class of satellite user, reflecting the sector of society using the satellite, or the sector a satellite is designed to serve.<br>E.g. A satellite is used by civil, commercial, governmental, military users, etc. |
| Purpose | Purpose | May also be represented by the class, Function. |
| Detailed Purpose | <sub-classes below in Fig.3> | Purpose hierarchy formed. Resolved with subclasses of Purpose class |
| Class of Orbit | Orbit<br><sub-classes below in Fig.2> | Orbit hierarchy formed. Resolved with Orbit class and subclasses |
| Type of Orbit | | Merged into sub-classes of Orbit. |
| Longitude of GEO (degrees) | Longitude_Of_GEO<br>Longitude_Of_GEO_value | A numerical value with unit of measure in degrees |
| Perigee (km) | Perigee<br>Perigee_value | A numerical value with unit of measure in Kilometer (km) |
| Apogee (km) | Apogee<br>Apogee_value | A class whose instances are particular apogees (of a specific numeric value for the spatial distance) of individual satellites.<br>A numerical value with unit of measure in Kilometer |
| Eccentricity | Orbital_Eccentricity<br><br>Orbital_Eccentricity_value | A class whose instances are particular eccentricities (of a specific numeric value) of individual satellites.<br>A decimal value less than or equal to 1 but not below 0. |
| Inclination (degrees) | Orbital_Inclination<br><br>Orbital_Inclination_value | A class whose instances are particular inclinations (of a specific value) of individual satellites.<br>Angular measure in degrees. |
| Period (minutes) | Orbital_Period<br><br>Orbital_Period_value | A class whose instances are particular periods (of a specific temporal duration) of individual satellites.<br>Time interval in minutes. |
| Launch Mass (kg.) | Launch_Mass | A numerical value with unit of measure in Kilogram (kg) |
| Dry Mass (kg.) | Dry_Mass | A numerical value with unit of measure in Kilogram |
| Power (watts) | Artificial_Satellite_Power | A numerical value with unit of measure in watts |
| Date of Launch | Launch_Date | Numerical date value |
| Expected Lifetime | Satellite_Expected_Lifetime | Time interval: numerical value with unit of measure in years |
| Contractor | Contractor | A social organization, agency, company, institution, etc. that |
| Country of Contractor | | Resolved into Contractor class and *has_Country_of_Origin* relation |
| Launch Site | Launch_Site | A socio-political site or geographic location, such as a city |
| Launch Vehicle | Launch_Vehicle | A type of vehicle used to help insert the satellite into orbit |
| COSPAR Number | COSPAR_Number | Alphanumeric string |
| NORAD Number | NORAD_Number | Alphanumeric string |
| Comments | Satellite_Comment | Alphanumeric string containing supplemental or descriptive information |

|  |  | about the satellite, e.g., satellite systems, components, functions, it's place in constellation, etc. |
|---|---|---|

**Table 2**: UCS Database terms and their corresponding ontology classes with description.

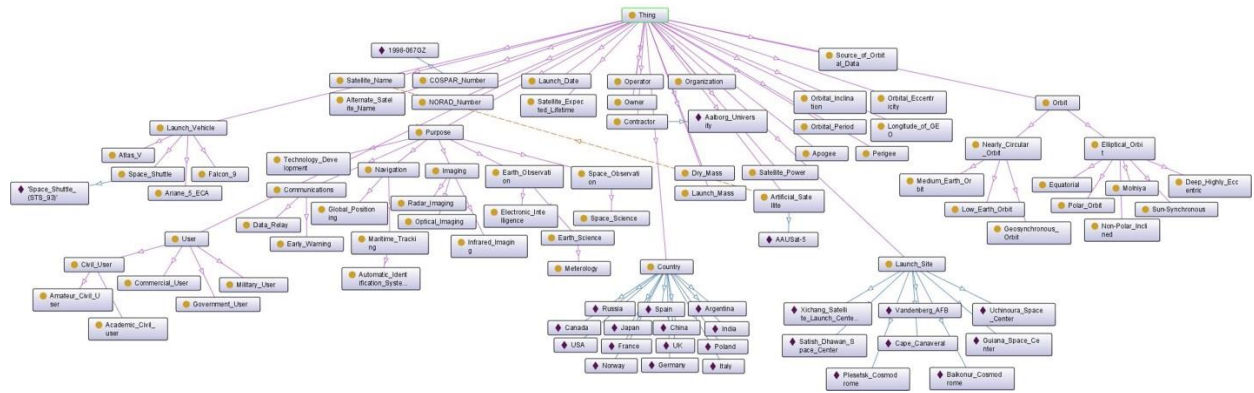

**Figure 1.** OntoGraf depiction of class (circles) hierarchy with instances (diamonds) using Protégé.

*2.1 Orbit Taxonomy & Orbital Properties*

The UCSSD distinguishes between Class of orbit and Type of orbit. Generally speaking there are various ways to categorize orbits, but I have not kept this particular distinction. Rather, I created a single **Orbit** hierarchy (figure 2) by merging the respective sub-classes and sub-types according to the description in the UCS manual. The hierarchy reflects only those orbits mentioned in the UCSSD.[5] If the classification proves to be insufficient, i.e., if the orbit class vs. type distinction is needed, the respective classes and relations can be added to the ontology. The primary orbital feature differentiating the two main orbit categories—**Elliptical Orbit** and **Nearly Circular Orbit**—is orbital eccentricity. It describes the shape of an orbit, and is one of the Keplerian Orbital Elements or Parameters, here subsumed as a type of **Orbital_Property**.

---

[5] As such, drift orbits and orbits at Lagrange points are not included, orbits that must be (and are) represented in a more complete domain ontology (e.g., the SSAO[5]).

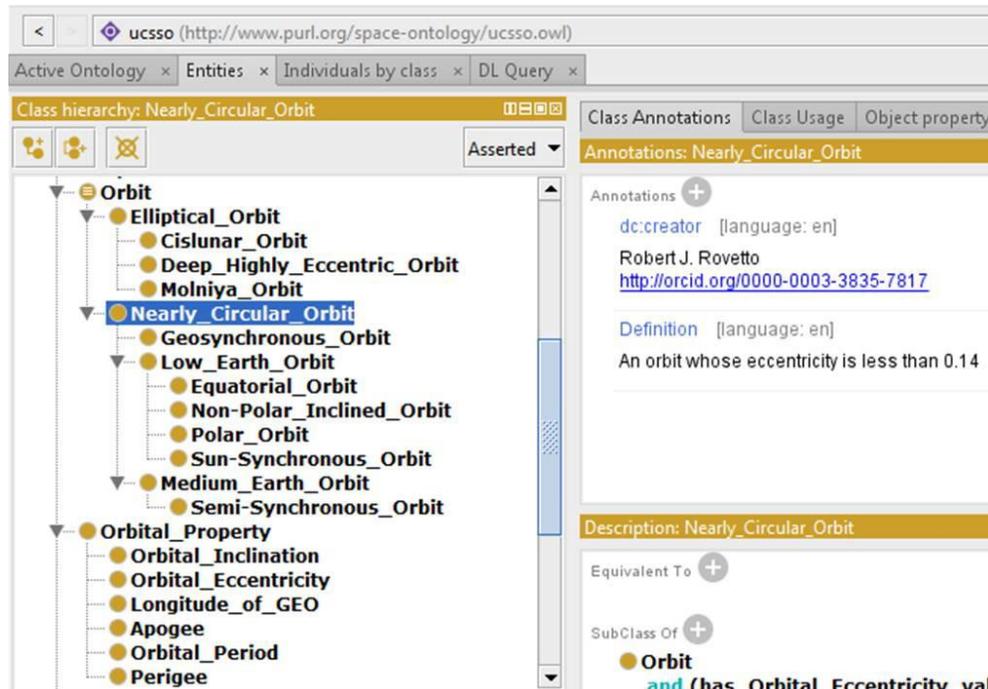

**Figure 2.** Screenshot of Orbit and Orbital Property classes of UCSSO, displayed in Protégé ontology editor.

Two slightly different ways to model the relationship between orbits, satellites and orbital parameters are as follows. First, I assert classes for each orbital parameter (e.g. **Orbital Eccentricity**). Their instances are added, representing each particular parameter for an individual satellite (e.g., AAUSat-4_Orbital Eccentricity). I also add classes of satellites based on their function as described in UCSSD (e.g., **Earth-Observing_Satellite**) with instances representing each individual satellite (e.g., AAUSat4). Satellite and parameter instances are related via orbital parameter Object Property relations (e.g., *has_Orbital_Eccentricity*) (listed in section 2.4). Orbital parameter instances are then related to a numeric value via a Data Property relation (e.g. *has_Orbital_Eccentricity_value*).[6] This yields a formal expression such as:

AAUSat-4 *has_Orbital_Eccentricity* AAUSat-4 Orbital Eccentricity *has_Orbital_Eccentricity_value* 0.02     (1)

According to the second approach the orbital properties—eccentricity, inclination, perigee, apogee, longitude of GEO, and period—are modeled merely with Protégé Data Properties (binary relations) whose range is again a numeric value. It does not use orbital parameter classes, thereby shortcutting the relationship to orbital parameter instances. A formal express reads: AAUSat-4 *has_Orbital_Eccentricity_value* 0.02. This is simpler, and may have a slight edge in terms of computationally performance (e.g., with automated reasoners), but is arguably not as ontologically accurate, semantically complete, or conceptually precise. Further research is in order.

In any case, general knowledge, rules and restrictions are necessary to formalize at the class level. For example, the fundamental domain knowledge that all (closed) orbits have an orbital eccentricity, inclination, period, perigee, apogee, etc., is formalized. Similarly, given that eccentricities are only from 0 to 1 for all orbits (excluding parabolic and hyperbolic trajectories), the Data Property has_Orbital_Eccentricity_value has a value restriction accordingly.

The UCSSD classification restricts the eccentricity of Nearly Circular Orbits to no greater than 0.14. An approximate but more expressive formalization using Firstorder predicate logic (FOL) is as follows. Note that the instance_of relation is a domain-neutral formal ontological relation relating categories to their instances. '∀' is the universal quantifier ("Every", "Each", "For all"), '∃' the existential quantifier ("there exists"), '∧' conjunction ("and") ,'→' the conditional ("if... then"), and lowercase letters are instances.

---

[6] I do likewise for their orbits: rather than relating the satellite to the parameters, one can relate them to orbits.

UCSSO:Nearly_Circular_Orbit =def.  (2)
∀x [instance_of(x, Nearly_Circular_Orbit) →
instance_of(x, Orbit) ∧ ∃y[has_Orbital_Eccentricty_value(x,y) ∧ y≤0.14]]

According to the first modeling approach, but with the **Orbit** class, a FOL formalization is:

∀x [instance_of(x, Nearly_Circular_Orbit) → instance_of(x, Orbit) ∧  (3)
∃y,z [has_Orbital_Eccentricty(x,y) ∧ instance_of(y, Orbital_Eccentricity)
∧ has_Orbital_Eccentricty_value(y,z) ∧ z≤0.14]]

From an ontological perspective, the first approach is ontologically richer, describing more entities in the universe of discourse whereas the former is ontologically sparser. I have questions as to the ontological status of many of these entities, questions to be investigated, and so I include both strategies tentatively. Note that the time (Epoch) element is implicit, but necessary in future development since the orbital parameters change over time.

*2.2 Purpose Taxonomy* – **Functions of Satellites**

The UCS manual describes Purpose (of satellites) as "The discipline in which the satellite is used in broad categories". Based on this description and that of the Detailed Purpose field, the latter are subclasses of the former. From a philosophical perspective, however, there are conceptual and ontological differences between a discipline and a purpose. We can just as easily assert a binary relation, *has_Discipline_of_Investigation* whose domain and range are **Artificial_Satellite** and **Discipline** (or instances thereof). UCSSO includes the following hierarchy (figure 3).

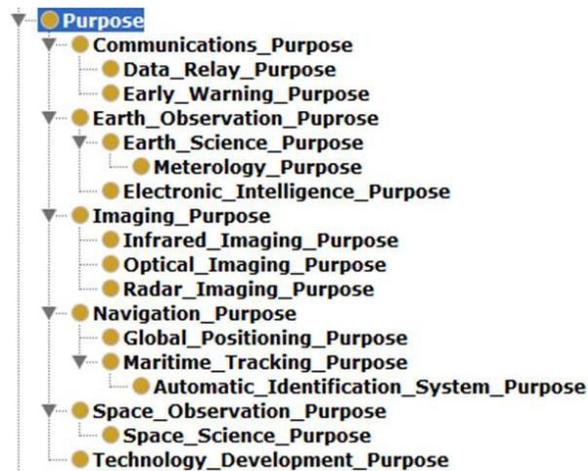

**Figure 3.** Taxonomy of Purpose classes of UCSSO, displayed in Protégé ontology editor.

A similar concept to that of purpose is function. Function is an ontological category often ascribed of engineering artifacts, making it attributable to artificial satellites. The literature is replete with analyses of functions and much can be said on both concepts, but I will not delve into it here. For this communication, the reader may use Purpose and Function interchangeably, e.g., reading 'Communications_Purpose' as 'Communications_Function'.

*2.3 Social and Political Entity Terms*

Social entity classes such as **User** (and its subclasses, e.g., **Civil_User**), **Owner**, and **Operator** may be organizations such as companies, space agencies, satellite operators from industry, military departments, universities and academic departments with satellite databases. The UCSSD mentions particular users and owners, such as: the University of Aalborg, Asia Broadcast Satellite Ltd., US Air Force, Aerospace Corporation, and the European Space Agency. These individuals are asserted in the ontology as instances of their respective classes: **Company**, **University**, **Space Agency**, etc. Likewise for **Country** and **Organization**, which I assert as distinct classes because of their ontological differences. If desired, these social and political entity classes can be imported from or mapped to an existing resource that models organizational entities.

Very briefly, the upper-level ontological category of **Role** may be helpful to characterize Owner and Operator, because a company or agency can be an owner or operator of an artificial satellite at one time but not another. Roles are entities that a role-holder plays over a period time in certain states of affairs. They are non-rigid properties according to [27]. That is, arguably nothing is necessarily an Owner or Operator, but only so under certain conditions.

*2.4 Relations & Supplemental Upper-level Classes*

Table 3 lists the binary relational predicates (Object and Data Properties) to relate the foregoing classes and instances. This allows us to express different characteristics associated with satellites. They are presented as binary because OWL (typically used in Protégé) is unfortunately limited to binary predicates, one expressive limitation of the ontology language. In reality, such relations will more accurately have more than two arguments.

| Relation | Domain (class) | Range |
|---|---|---|
| has_Orbit / has_Orbit_type | Artificial_Satellite | Orbit |
| has_Country_of_Origin | Artificial_Satellite | Country |
| is_registered_Country_in_UN_Register_of_Space_Objects_for | Country | Artificial_Satellite |
| is_registered_Organization_in_UN_Register_of_Space_Objects_for | Organization | Artificial_Satellite |
| has_Operator | Artificial_Satellite | Operator |
| has_Owner | Artificial_Satellite | Owner |
| has_User | Artificial_Satellite | User |
| has_Contractor | Artificial_Satellite | Contractor |
| has_Identifier | Artificial_Satellite | |
|     has_COSPAR number | | COSPAR_Number |
|     has_NORAD number | | NORAD_Number (Alphanumeric values) |
| has_Purpose | | Purpose |
| has_Function | | Function |
| has_Orbital_Property / has_Orbital_Parameter | Artificial_Satellite, Orbit | |
|     has_Orbital_Inclination | | Orbital_Inclination |
|     has_Orbital_Inclination_value | | A numeric value. |
|     has_Orbital_Eccentricity | | Orbital_Eccentricity |
|     has_Orbital_Eccentricity_value | | A numeric value. |
|     has_Logitutde_of_GEO | | Logitutde_of_GEO |
|     has_Logitutde_of_GEO_value | | Degrees. |
|     has_Perigee | | Perigee |
|     has_Perigee_value | | A numeric value, km. |
|     has_Apogee | | Apogee |
|     has_Apogee_value | | A numeric value, km. |
| has_Dry_Mass | Artificial_Satellite | Kilograms |
| has_Launch_Mass | Artificial_Satellite | Kilograms |
| has_Power_value | | Watts |
| has_Date_of_Launch | Artificial_Satellite, Launch_Vehicle | Date |
| has_Expected_Lifetime | Artificial_Satellite, Launch_Vehicle | Time Interval/Period (e.g. years) |

| has_Launch_Site | Artificial_Satellite | Launch_Site |
| has_Launch_Vehicle | Artificial_Satellite | Launch_Vehicle |

**Table 3.** Relational predicates, and their domain and range.

For example, as with describing orbital properties, measurements, and social/political aspects such as ownership, temporal indexing is often necessary. A time, **t**, represents temporal moments, and a range, [t, tn] represents temporal intervals. These can serve as a third argument, something easily expressible in FOL, higher-order logics, and the CLIF ontological language. Computational complexity notwithstanding, for this reason, n-ary relations where n≥2, should be sought after if one aims for a complete and detailed ontological representation.

The UCSSD includes various numeric figures, e.g., for the orbital parameters. As mentioned in section 2.1, one strategy to represent these entities is to assert Data Properties in Protégé whose Range is a numeric *value*. Accordingly, the word 'value' is included in the relation name to disambiguate from Object Property relations. Finally, figure 4 portrays UCSSO at the class level, where ovals, grey boxes, and arrows signify classes, groupings of classes, and relations, respectively.

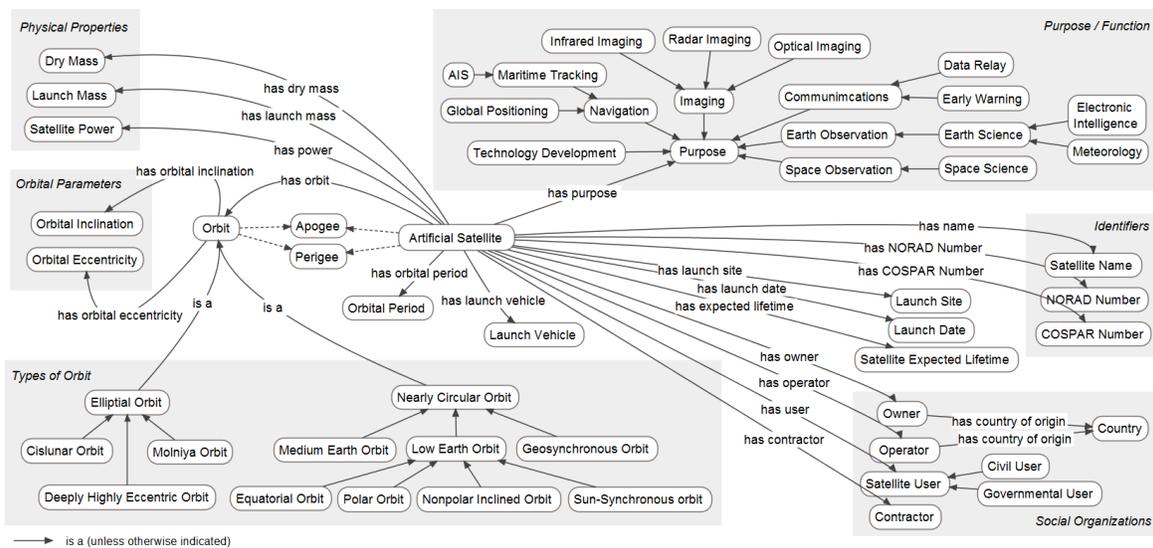

**Figure 4**. The UCS Satellite Ontology. Grey boxes signify groupings of classes.

Since UCSSO presently reflects only those concepts in the UCSSD, a more complete representation of the domain would call for using a domain ontology such as the SSAO [5].

## 3. A Domain Ontology for Satellite Databases: The SSA Ontology and The Orbital Space Domain Ontology Project

The preceding presented an ontology specifically of the Union of Concerned Scientists Satellite Database (UCSSO). This section discusses a domain reference ontology effort by the author that may be used by any satellite database, including the UCSSD. In all, this work is part of the orbital space domain ontology project [6][8], conceived in early form in 2011 and first published in [4](Rovetto 2015). The main production goal is one or more ontologies for astronautics, orbital debris, SSA and the space domain. These ontologies adopt the open world assumption, are currently under development, subject to revision, and open to cooperative development and partnerships. Presently, they are formalized in OWL. OWL files are available through the author, or in the near future at the persistent URL links in footnotes 1-3.

The SSA Ontology (SSAO) (Fig. 5) [5], for example, is a reference ontology for the SSA domain. It captures common knowledge and general concepts shared across the SSA and satellite community. A example user of the SSAO is an ontology-driven orrery project (in progress) by Daniel A. O'Neil at NASA Marshal Space Flight Center [28]. The SSAO contains formally defined category and relation terms necessary to annotate SSA data, while expressing a holistic real-world representation of the domain. Thus, given the domoain of interest, the terms in UCSSO are terms in (and drawn from) the SSAO, as indicated by Figure 5.

This affords at least two options for curators of space object catalogs seeking to apply ontology to their information systems. One, locally-developed ontologies (e.g. UCSSO) can be mapped to the SSAO, import SSAO classes, or extend the SSAO.

Two, rather than developing a local ontology, each space actor can use the SSAO as their domain ontology. As a domain ontology, it is intended to offer a semantically-rich backbone vocabulary for any satellite database, whether of operational satellites (as in the UCSSD) or all orbital space objects and events (as in the U.S. Space Object Catalog). Although each database may use a unique data element or term for the same satellite aspect, annotating them with a higher-level formally-defined class from the SSAO will reduce ambiguity and add meaningful content to the data. It should also give each actor the option of sharing information. The SSAO may therefore serve as a potential, if indirect, link between federated databases, offering a common semantics and domain model. Table 4, revised from Table 1, expresses some central features of the SSAO, drawing in part on findings by (Raskin and Pan 2003). In either case, cooperative engagement with the SSAO will not only facilitate mappings between terms in each ontology, but also help develop the SSAO into a thorough domain ontology that space actors can use.

| | | | |
|---|---|---|---|
| **Common Space Domain Vocabulary:** | Universal/general SSA concepts expressed by categories and relation terms | | |
| **A Structured Terminology:** | Class-subsumption & other relations between categories, e.g., parthood, etc. | | |
| **Formally-defined SSA Terms:** | Natural language definitions (Human readable) | Formal definitions (OWL, FOL, CLIF). (Machine readable) | Rules, axioms, etc. |
| **A SSA Domain Model:** | Formalizes common knowledge of the satellite domain | Expresses meaning. Represents real-world orbital phenomena. | |
| **Reusability & Application-neutral:** | Common domain knowledge makes it application-neutral. | Heterogeneous space object databases can use it as a domain terminology/representation. They can also import it (or selected terms) into their own ontologies. | |
| **Scalability & Editable:** | Open world assumption. New terms can be added. Can be extended by more specialized ontologies. | | |
| **Facilitates Analysis:** | Query orbital information/data/ontology (e.g. SPARQL, SWRL, DL Query, etc.) | | |

**Table 4.** Central features of the SSA ontology & related ontologies (Rovetto, 2015), (Rovetto & Kelso, 2016), (Rovetto, 2016)

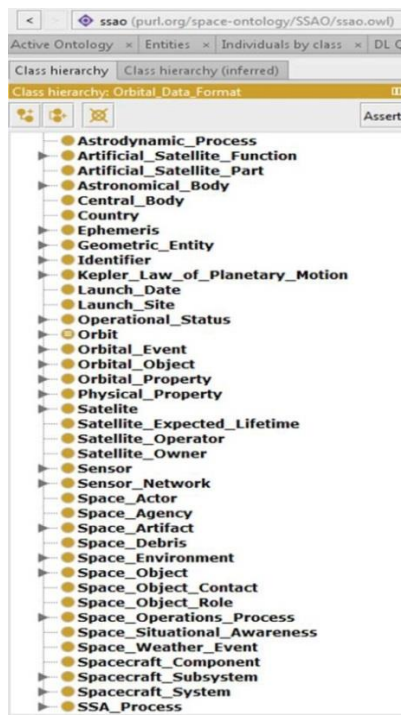

**Figure 5**. A portion of the Space Situational Awareness Ontology (SSAO) displayed in Protégé.

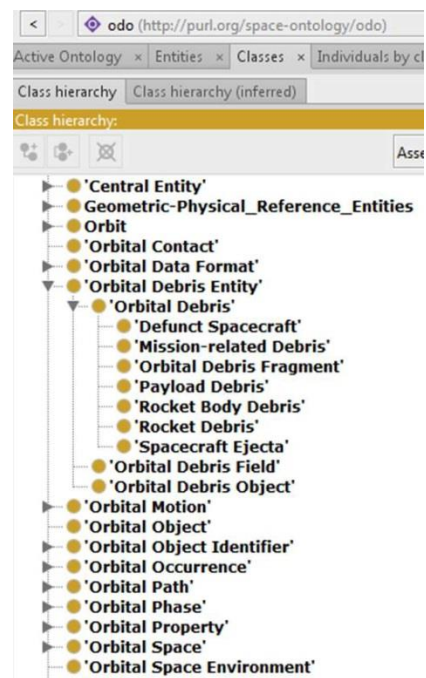

**Figure 6**. A portion of the Orbital Debris Ontology (ODO) displayed in Protégé ontology editor.

While the UCSSO focuses on active artificial satellites, a thorough domain reference ontology or ontology suite should include other sorts of orbital objects, e.g., inactive satellites and orbital debris, their interactions and relationships to eachother and to other relevant entities, etc. The overall project aims to formally represent all these entities via one or more modular ontologies, e.g., the SSAO.

Another ontology under development is the Orbital Debris Ontology7 (ODO) (Rovetto 2015, 2016d) (Fig. 6), which initiated the project concept with ODO's conception in 2011 and first publication in 2015. The minimum scope of ODO is orbital debris. That is, it minimally represents orbital debris objects and their physical properties, their interrelationships and the relations to other space objects and SSA entities. ODO classes can be used to annotate data on particular space debris objects, measurement data, etc. A subset of fundamental orbital terms in this project will form a generic Orbital Ontology (or 'orbitology' ontology) for use by ODO, the SSAO, other ontologies for space, and applications. UCSSO includes some of these terms.

Not only do they contain equivalent classes to those concepts expressed in the UCSSD, making part of the SSAO importable into a satellite ontology (e.g. UCSSO), but these ontologies also have more general domain-specific classes capable of subsuming them. Example include: Satellite, Spacecraft, Space Artifact, Space Object, Orbital Element, Orbital Property, Central Body, Orbital Path, Spacecraft Maneuver, etc. This helps make the ontologies scalable, as well as application- independent: more specialized astro ontologies can extend them or reuse selected terms. The SSAO can like extend ontologies or reuse terms. The SSAO and ODO can, for instance, be used with ontologies such as the NASA SWEET ontologies (SWEET) (Raskin and Pan 2003). Orbital and other properties are broadly categorized using more abstract classes such as Physical Property. At the most abstract levels of ontology engineering, these project ontologies (SSAO, ODO, etc.) are usable with foundational or top-level ontologies (Herre et al. 2006; Guizzardi and Wagner 2010; Mizoguchi 2010; DOLCE 2006), which provide the most general categories such as Property, Event, and Object.

## 4. Considerations, Applications and Future Work

Without expanding the coverage of UCSSO, i.e. adding more classes, relations and axioms, there will be domain entities that it does not represent. UCSSO is therefore minimally applicable to the UCSSD, the domain coverage being more-or-less limited to the concepts expressed in the UCS database. This is acceptable for satellite database administrators that need not use the ontology elsewhere. By contrast, the reference ontologies under development—the SSAO, ODO, a generic orbital ontology, etc.—provide a more thorough coverage of the domain, allowing use across information systems.

These domain reference ontologies can be applied not only to space object catalogs, but to space environment visualizations. For example, (Quartz 2015; Grego 2014)[24][25] uses the UCSSD to graphically visualize satellites. Similarly, and as expressed in NASA Datanaut presentations (part of the NASA open data initiative) [28] uses the SSAO to populate orbital information toward visualizing an interactive animation of the solar system. Thus, there is potential for pedagogical innovation, e.g., ontology-based visual aids to learning for astronomy and astronautics. Finally, given overlapping interest in taxonomy development, potential application of these ontologies may be to the taxonomy concepts found in publications such as [34] (Fruh et al. 2013).

Future work includes development, testing and implementation: class definitions, taxonomy organization, formalizing domain knowledge, subject-matter research, ontology engineering research, data querying, etc. Participation from subject-matter experts in space and data/computer science disciplines is encouraged for thorough development. Satellite observation and tracking involves uncertainty and prediction both in and outside of astrodnyamic models: orbit propagation, collision estimation, orbital debris origins, space object identification and tracking, etc. Ontological treatments of uncertainty, causality and predictive processes will therefore be helpful. A philosophically-rigorous approach, if employed thoroughly, will help to clarify concepts, introduce helpful distinctions, and precisely characterize domain entities in a formal and platform independent manner.

**5. Conclusion**

This paper developed an ontology for the Union of Concerned Scientists (UCS) Satellite Database. In doing so, I demonstrated that the Space Situational Awareness Ontology (SSAO), an existing domain ontology for the satellite and SSA community, can be used either in conjunction with local ontologies like UCSSO, or as an alternative, i.e., as the domain reference ontology for various satellite databases.

This ongoing effort is intended to provide a common domain-specific computable terminology and knowledge model for space data systems. Where data is drawn from multiple sensors or databases, ontologies should foster information fusion via this backbone terminology. These ontologies may also stimulate data exchange, retrieval and search across federated databases, as well as offer ontological classifications of space objects and astrodynamic phenomena.

**Acknowledgements.**

I am grateful for the reviews from the journal reviewers. Thanks also go to Mathias Brochhausen of the University of Arkansas, and Alan Ruttenberg of the State University of New York at Buffalo for their helpful input and their time in answering my questions. Thanks to Alan for the modeling suggestion mentioned in the 'Orbit taxonomy and orbital properties_ subsection.

**References**

[1] Union of Concerned Scientists (UCS) Satellite Database. URL= http://www.ucsusa.org/nuclear-weapons/space-weapons/satellite-database
[2] "UCS Common Misconceptions". URL= https://s3.amazonaws.com/ucs-documents/nuclear-weapons/sat-database/common-misconceptions.pdf

[3] UCS Database (Excel format) URL= https://s3.amazonaws.com/ucs-documents/nuclear-weapons/sat-database/2-25-16+update/UCS_Satellite_Database_1-1-16.xls

[4] R.J. Rovetto, An Ontological Architecture for Orbital Debris Data, *Earth Science Informatics* **9**(1) (2015), 67–82. DOI: 10.1007/s12145-015-0233-3. URL= http://link.springer.com/article/10.1007/s12145-015-0233-3

[5] R.J. Rovetto, T.S. Kelso, Preliminaries of a Space Situational Awareness Ontology. Presented at 26[th] AIAA/AAS Space Flight Mechanics Meeting, Napa, California, USA, Feb 14-18th, 2016. Proceedings published in *Advances in the Astronautical Sciences,* vol. 158, Univelt Inc.
Univelt paper URL= http://www.univelt.com/book=5920. Preprint URL= https://arxiv.org/ftp/arxiv/papers/1606/1606.01924.pdf

[6] R.J. Rovetto. The Orbital Space Environment and Space Situational Awareness Domain Ontology – Towards an International Information System for Space Data. The Advanced Maui Optical and Space Surveillance Technologies (AMOS) Conference, Sept.20-23, 2016, Maui, Hi, USA. URL= http://www.amostech.com/TechnicalPapers/2016/Poster/Rovetto.pdf

[7] "Three Reasons the UCS Satellite Database is Different from Other Satellite Catalogs". URL=http://allthingsnuclear.org/lgrego/three-reasons-the-ucs-satellite-database-is-different-from-other-satellite-catalogs

[8] R.J. Rovetto "Orbital Space Environment and Space Situational Awareness Domain Ontology" In CEUR workshop proceedings for The Joint Ontology Workshops, at the 9th International Conference of Formal Ontology for Information Systems (FOIS) Early Career Symposium, Annecy, France July 2016. Project summary paper, and poster session.

[9] R.J. Rovetto "Ontology Architectures for the Orbital Space Environment and Space Situational Awareness Domain" International Workshop on Ontology Modularity, Contextuality, and Evolution at the 9th International Conference on Formal Ontology in Information Systems, Annecy France, July 2016.

[10] R.J. Rovetto "Orbital Debris Ontology", Center for Orbital Debris Education and Research (CODER), University of Maryland, 15-17 November 2016.

[11] R.J. Rovetto, Foundations of Space Object Ontology (Forthcoming).

[12] Quantities, Dimensions, Units and Data Types Ontology (QUDT) URL= http://www.qudt.org/

[13] Staab S, Studer R (Ed.)(2009) Handbook on Ontologies, International Handbooks on Information Systems, Springer 2nd ed.

[14] M.Uschold, "Ontologies and Database Schema: What's the Difference?" (Powerpoint Presentation) URL= https://pdfs.semanticscholar.org/b44f/a4592b69183c1965d0075dea1a3bc58dfbfe.pdf

[15] "What Is an Ontology?", Guarino, Nicola, Oberle, Daniel, and Staab, Steffen, in S. Staab and R. Studer (eds.), Handbook on Ontologies, International Handbooks on Information Systems, Springer-Verlag Berlin Heidelberg 2009. doi: 10.1007/978-3-540-92673-3

[16] Rudi Studer, V. Richard Benjamins, Dieter Fensel. Knowledge Engineering: Principles and methods. Data & Knowledge Engineering, Vol.25 (1-2), pp.161-197. March, 1998. DOI: http://dx.doi.org/10.1016/S0169-023X(97)00056-6. URL= http://www.sciencedirect.com/science/article/pii/S0169023X97000566

[17] Borne K.D. (2010) Astroinformatics: data-oriented astronomy research and education. Earth Science Informatics, Volume 3, Issue 1, pp 5-17. URL= http://link.springer.com/article/10.1007%2Fs12145-010-0055-2

[18] National Aeronautical and Space Administration (NASA) Semantic Web for Earth and Environmental Terminology (SWEET) Ontologies, Jet Propulsion Laboratory, California Institute of Technology. URL= http://sweet.jpl.nasa.gov/

[19] Common Logic - International Organization for Standards. URL= http://www.iso.org/iso/catalogue_detail.htm?csnumber=39175

[20] Web Ontology Language (OWL) URL= https://www.w3.org/TR/owl-features/

[21] Ian Horrowicks. Ontologies and Databases (Powerpoint presentation). Information Systems Group, Oxford University Computing Laboratory. URL= www.cs.ox.ac.uk/ian.horrocks/Seminars/download/onto-db.ppt

[22] UCS Satellite Database User's Manual 1-1-16. URL= https://s3.amazonaws.com/ucs-documents/nuclear-weapons/sat-database/2-25-16+update/User+Guide+1-1-16+wAppendix.pdf

[23] Protégé ontology editor. URL= http://protege.stanford.edu/

[24] "The World Above Us: This is every active satellite orbiting the earth", Quartz, URL= http://qz.com/296941/interactive-graphic-every-active-satellite-orbiting-earth/

[25] "Visualizing the UCS Satellite Database" L.Grego, URL=http://allthingsnuclear.org/lgrego/visualizing-the-ucs-satellite-database

[26] "Exploring and cleaning the Union of Concerned Scientists database of Earth Satellites" BaysDB project, The MIT Probabilistic Computing Project URL= http://probcomp.csail.mit.edu/bayesdb/satellites-notebook.html

[27] N.Guarino and C.Welty. "An Overview of OntoClean", in Handbook on Ontologies, International Handbooks on Information Systems pp. 151-171, Springer Berlin Heidelberg, 2004. DOI: 10.1007/978-3-540-24750-0_8.

[28] Daniel O'Neil, Ontology-driven Orrery, NASA , Marshal Space Flight Center.
URL= https://github.com/daoneil/spacemission/tree/master/OntologyDrivenOrrery
Orrery URL= http://daoneil.github.io/spacemission/OntologyDrivenOrrery/An_Orrery_in_ThreeJS.html

[29] General Formal Ontology (GFO) - A Foundational Ontology Integrating Objects and Processes Heinrich Herre, Barbara Heller†, Patryk Burek, Robert Hoehndorf, Frank Loebe, Hannes Michalek, Part I: Basic Principles, Version 1.0.1. URL= http://www.onto-med.de/ontologies/gfo/

[30] Giancarlo Guizzardi, Gerd Wagner. Using the Unified Foundational Ontology (UFO) as a Foundation for General Conceptual Modeling Language, Theory and Applications of Ontology: Computer Applications, pp 175-196, 12 August 2010, Springer Netherlands. doi: 10.1007/978-90-481-8847-5_8

[31] Yet Another More Advanced Top Level Ontology (YAMATO), URL= http://download.hozo.jp/onto_library/upperOnto.htm

[32] Descriptive Ontology for Linguistic and Cognitive Engineering (DOLCE) URL= http://www.loa.istc.cnr.it/old/DOLCE.html

[33] Raskin R, Pan M (2003) Semantic Web for Earth and Environmental Terminology. CEUR Workshop Proceedings Vol-83. ISSN 1613-0073. Proceedings of workshop on Semantic Web Technologies for Searching and Retrieving Scientific Data. URL= http://ceur-ws.org/Vol-83/sia_7.pdf

[34] Fruh C, Jah M, Valdez E, Kervin P, Kelecy T (2013) Taxonomy and Classification Scheme for Artificial Space Objects. In Proceedings of Advanced Maui Optical Space Surveillance Conference, Maui, USA